\documentclass{article}

\usepackage{microtype}
\usepackage{graphicx}
\usepackage{subfigure}
\usepackage{subcaption}
\usepackage{booktabs} 

\usepackage{hyperref}
\usepackage{multirow}
\usepackage{spverbatim}

\usepackage{tcolorbox}
\newtcolorbox{promptbox}[1][]{
    colback=gray!5!white, colframe=gray!50!black,
    fontupper=\small\ttfamily, 
    arc=1mm, boxrule=0.5pt,
    title=#1,
    left=6pt, right=6pt, top=6pt, bottom=6pt
}

\usepackage[accepted]{icml2025}

\usepackage{amsmath}
\usepackage{amssymb}
\usepackage{mathtools}
\usepackage{amsthm}

\usepackage[capitalize,noabbrev]{cleveref}

\theoremstyle{plain}

\theoremstyle{definition}

\theoremstyle{remark}

\usepackage[textsize=tiny]{todonotes}

\usepackage{xcolor}

\icmltitlerunning{Self Knowledge Re-expression: A Fully Local Method for Adapting LLMs to Tasks Using Intrinsic Knowledge}

\begin{document}

\twocolumn[
\icmltitle{Self Knowledge Re-expression: A Fully Local Method\\ for Adapting LLMs to Tasks Using Intrinsic Knowledge}




\begin{icmlauthorlist}
\icmlauthor{Mengyu Wang\textsuperscript{*}}{uni,comp}
\icmlauthor{Xiaoying Zhi}{comp}
\icmlauthor{Zhiyi Li}{comp}
\icmlauthor{Robin Schmucker}{comp}
\icmlauthor{Shay B. Cohen}{uni}
\icmlauthor{Tiejun Ma}{uni}
\icmlauthor{Fran Silavong}{comp}
\end{icmlauthorlist}

\icmlaffiliation{uni}{The University of Edinburgh}
\icmlaffiliation{comp}{JPMorgan Chase \& Co}


\icmlcorrespondingauthor{Mengyu Wang}{mengyu.wang@ed.ac.uk}

\icmlkeywords{Machine Learning, ICML}
\vskip 0.3in
]

\newcommand{\shaycomment}[1]{\textcolor{blue}{#1 -- Shay}}
\newcommand{\mengyucomment}[1]{\textcolor{red}{#1 -- Mengyu}}

\printAffiliationsAndNotice{} 

\newenvironment{enumeratesquish}[2]{\begin{list}{\labelenumi}{\setlength{\itemsep}{#1}\setlength{\labelwidth}{#2}\setlength{\leftmargin}{\labelwidth}\addtolength{\leftmargin}{\labelsep}}}{\end{list}}

\newcommand{\citationnote}[1]{\textcolor{blue}{#1 --CitationNote}}
\newcommand{\mengyu}[1]{\textcolor{blue}{#1 --MengyuNote}}
\newcommand{\robin}[1]{\textcolor{teal}{#1 -- Robin}}
\newcommand{\zhiyi}[1]{\textcolor{orange}{#1 -- Zhiyi}}
\newcommand{\xiaoying}[1]{\textcolor{pink}{#1 -- Xiaoying}}

\let\thefootnote\relax\footnotetext{
\textsuperscript{*}Work completed during an internship at the Machine Learning Center of Excellence (MLCOE), JPMorgan Chase \& Co.


}

\begin{abstract}
While the next-token prediction (NTP) paradigm enables large language models (LLMs) to express their intrinsic knowledge, its sequential nature constrains performance on specialized, non‑generative tasks.
We attribute this performance bottleneck to the LLMs' knowledge expression mechanism, rather than to deficiencies in knowledge acquisition.
To address this, we propose Self‑Knowledge Re‑expression (SKR), a novel, task‑agnostic adaptation method.
SKR transforms the LLM's output from generic token generation to highly efficient, task-specific expression. SKR is a fully local method that uses only unannotated data, requiring neither human supervision nor model distillation.
Experiments on a large financial document dataset demonstrate substantial improvements: over 40\% in Recall@1 for information retrieval tasks, over 76\% reduction in object detection latency, and over 33\% increase in anomaly detection AUPRC. 
Our results on the MMDocRAG dataset surpass those of leading retrieval models by at least 12.6\%.
\end{abstract}

\section{Introduction}
\label{sec:introduction}
Large language models (LLMs) have fundamentally transformed natural language processing (NLP), evolving from simple token generators into vast knowledge repositories~\cite{radford2018improving,radford2019language,brown2020language}. Traditionally, LLM performance is viewed as being directly related to the volume and quality of knowledge encapsulated within their parameters, acquired during the pre-training phase~\cite{kaplan2020scaling}. This perspective has driven continuous efforts to scale up model size and pre-training corpora, underscoring the belief that knowledge acquisition is the primary driver of LLMs' capability.

\begin{figure}[t]
\centering
\begin{center}
   \includegraphics[width=0.98\linewidth]{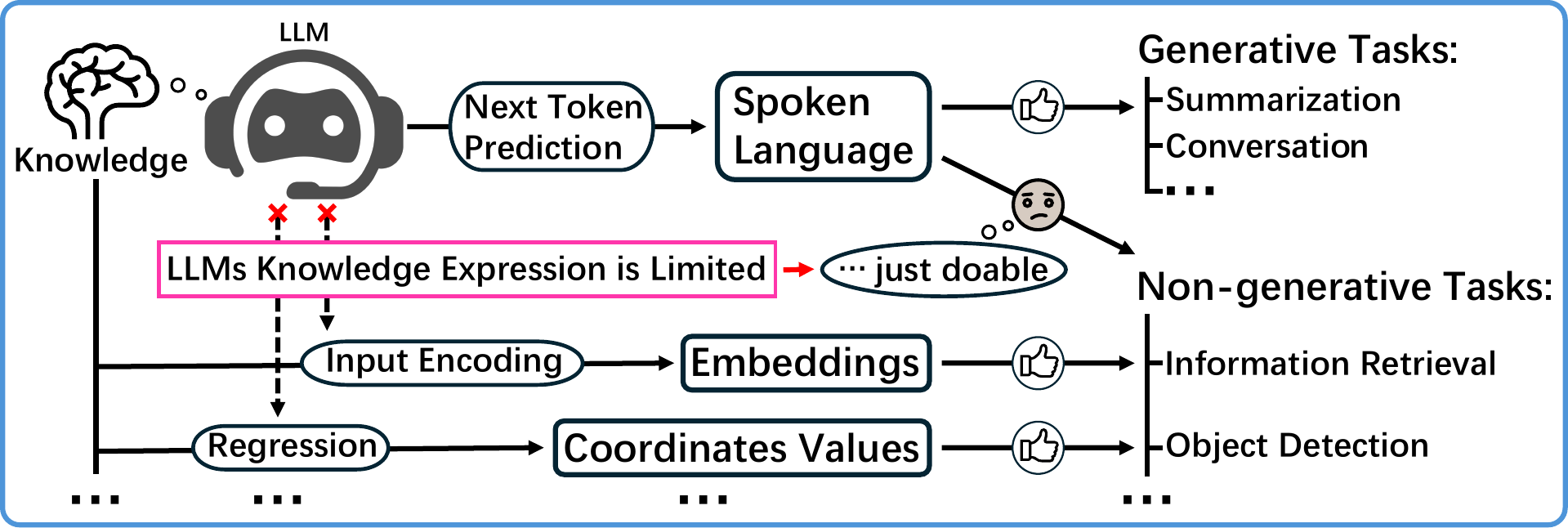}
    \vspace{-0.3cm}
   \caption{The inherent limitation of LLMs' knowledge expression: Next-token prediction serves as a universal paradigm but is a suboptimal output mechanism for many non-generative tasks.}
\label{fig:motivation}
\end{center}
\vspace{-0.4cm}
\end{figure}

However, knowledge acquisition is not the only performance bottleneck. Recent studies suggest that the intrinsic knowledge within LLM parameters is not merely a collection of statistical co-occurrences but is structured and highly reusable~\cite{rothenfusser2025vector}. Therefore, we argue that knowledge expression, the process of organizing latent parametric knowledge into coherent and effective task outputs, is a critical, yet under-explored, factor in model performance.
The richness of the knowledge existing within current LLMs implies that a promising direction for performance enhancement lies in optimizing their \textit{knowledge expression} rather than \textit{knowledge acquisition}.

Currently, the prevailing mechanism for LLMs' knowledge expression is the next-token prediction (NTP) paradigm. While NTP facilitates the generation of fluent text, discrete, token-based strings are often ill-suited for tasks requiring non-textual outputs, such as high-quality embeddings or continuous values. In these contexts, the reliance on NTP restricts a model’s ability to articulate its internal knowledge, necessitating cumbersome post-processing to bridge the gap between linguistic outputs and task requirements~\cite{ma2024unifying,lukasik2025better}. Consequently, as illustrated in Figure~\ref{fig:motivation}, while NTP suits most generative tasks, it imposes constraints on many non-generative tasks.

The limitations of the NTP paradigm are manifold, primarily stemming from a fundamental objective mismatch. In information retrieval tasks, for instance, the ideal expression of knowledge consists of dense, high-dimensional vectors optimized for semantic proximity~\cite{karpukhin2020dense}. However, embeddings derived from NTP-optimized models often fail to meet these requirements~\cite{ma2024unifying}. 
Furthermore, the sequential nature of NTP struggles with global planning and "lookahead" tasks~\cite{bachmann2024pitfalls}. In tasks such as \textit{object detection} or \textit{numerical reasoning}, a numerical value (e.g., a coordinate) is split into multiple tokens. Errors in early token predictions accumulate, compromising both numerical accuracy and structural integrity.
Additionally, NTP imposes further constraints, such as the absence of task-oriented probability distributions for classification and the inefficiency of post-processing purely linguistic outputs. These limitations underscore a fundamental problem: although NTP offers a universal interface by mapping all information to a textual format, it is rarely the optimal expression for specific tasks.

We propose addressing these challenges by transitioning the expression mechanism ($E$) from the generic $E_{\textit{ntp}}$ to a task-specific expression $E_T$. Unlike standard supervised fine-tuning (SFT), which aims to refine model knowledge or instill domain nuances through extensive annotated data~\cite{hanparameter,zheng2024fine}, our method uses only unannotated data and the model's intrinsic knowledge. Therefore, this approach enables highly efficient task adaptation by bypassing the requirement of high-quality external supervision.

We term our method Self Knowledge Re-expression (SKR), a fully local process that unlocks the latent potential of LLMs by using their intrinsic parametric knowledge without external human or model-based supervision. This method consists of two steps, \textit{Self Annotation} and \textit{Knowledge Re-expression}. First, the LLM uses its generative capability ($E_{\textit{ntp}}$) to create its own training labels from unannotated data. The model is then fine-tuned to map its internal states directly to a task-specific output expression $E_T$. This process is significantly more cost-effective than SFT and mitigates data exposure risks, which is vital for sensitive sectors like finance and healthcare.

In summary, our contributions are as follows:

\textbf{1. Formalization of Knowledge Expression Bottleneck:} We are the first to rigorously decouple knowledge expression from LLMs' parametric knowledge. We define the expression mechanism $E$ as a trainable component and formalize the adaptation process as the transition $E_{\textit{ntp}} \to E_T$.
    
\textbf{2. The SKR Method:} We propose a novel, task-agnostic method that enables an LLM to adapt to specific tasks by fine-tuning itself to a task-suitable expression $E_T$ without using external supervision. As a fully local process, SKR offers an economical and secure route for task adaptation.
    
\textbf{3. Validation Across Diverse Task Types:} We demonstrate the versatility of SKR across three distinct task types that reveal NTP's limitations: information retrieval, object detection, and anomaly detection. Our results show that SKR-adapted models' consistently and significantly improve performance, demonstrating that optimizing expression can extract substantial latent potential from LLMs.

\section{Related Work}
\label{sec:related_work}
The evolution of LLMs has been largely defined by the scaling hypothesis, which posits that model performance improves predictably with increases in parameter count, training data and computation~\cite{hestness2017deep,kaplan2020scaling}. Subsequent research has emphasized that optimizing the volume and quality of training tokens to enhance knowledge acquisition is as critical as scaling model size~\cite{brown2020language,hoffmann2022training}. To further augment these capacities, retrieval-augmented generation (RAG) has emerged to externalize knowledge into non-parametric memory, improving performance on knowledge-intensive tasks~\cite{lewis2020retrieval}. However, despite these advancements in knowledge acquisition, LLMs still exhibit significant performance gaps in non-generative tasks~\cite{grabuloski2025enhancing,song2025injecting}. The prevailing strategy to bridge this gap remains fine-tuning models using task-specific datasets and reward signals~\cite{kaufmann2023survey,koukounas2024jina,xie2023pixiu}.

With increasing training data volume, model size, and risks associated with external data exposure, research has shifted toward unlocking the latent potential within pre-trained models without relying on extensive external supervision~\cite{chen2024self}. Current post-training paradigms have explored methods such as self-reflection~\cite{qiu2024training,yuan2024self}, weak supervision~\cite{yu2021fine}, and self-generated reward signals~\cite{shao2025spurious} to refine specific task performance. While these methods use a model’s internal parameters to enhance performance, they remain confined to the standard generative paradigm, focusing on optimizing the quality of next-token prediction (NTP).

However, a growing body of evidence suggests that the NTP paradigm is not a panacea for all tasks~\cite{bachmann2024pitfalls}. For non-generative applications, such as information retrieval or structured regression, task-oriented output formats have been found to outperform linguistic generation~\cite{wang2024mana,ma2024unifying,wang2024modeling}. Our work, SKR, is motivated by these insights and extends beyond mere annotation-free fine-tuning. Unlike existing self-improvement methods that refine what the model generates within the NTP paradigm, SKR fundamentally alters how the model expresses its knowledge. Using the model's native generative capability to extract its intrinsic knowledge and subsequently refining its expression mechanism to a task-optimized structure, SKR enables LLMs to achieve superior task performance and inference efficiency.

\section{Problem Setup}
\label{sec:problem_setup}

\subsection{LLMs' Knowledge Expression}
\label{sec:llms_knowledge_expression}
Completing an NLP task involves providing inputs to a model to elicit desired outputs. A model's ability to generate desired outputs stems from its parametric knowledge acquired during pre-training and stored within its parameters~\cite{wang2024knowledge}. In current LLMs, this knowledge is universally expressed through the next-token prediction (NTP) paradigm~\cite{achiam2023gpt,dubey2024llama}. 

Formally, given an LLM $M$ with parameters $\theta_K$ storing knowledge $K$. The standard NTP expression paradigm, $E_{\textit{ntp}}$, uses these parameters to generate tokens in sequence. This process is formalized as modeling the probability of the next token $w_n$ within a vocabulary $\mathcal{V}$, conditioned on the sequence of preceding tokens ($w_0^{n-1}$): 
\begin{equation} 
E_{\textit{ntp}}(w_0^{n-1}; \theta_K) \coloneqq { P(w_n = w \mid w_0^{n-1}; K) } \quad {w \in \mathcal{V}}. \end{equation}

For a given task $T \in \mathcal{T}$, where $\mathcal{T}$ denotes the set of all possible tasks, the execution of $T$ by LLM $M$ begins with a prompt with task-specific instructions and raw input data. Subsequently, the model $M$ applies the $E_{\textit{ntp}}$ paradigm to generate tokens as the task output. Within this pipeline, all task-specific characteristics and structural requirements are integrated into and inherently constrained by the sequential token stream.
Current research often treats $E_{\textit{ntp}}$ as a versatile, all-purpose expression mechanism, using it for both generative and non-generative tasks by encoding non-textual information (e.g., numerical values) into textual formats. Therefore, prior work has primarily focused on optimizing the parametric knowledge $K$ via fine-tuning or post-training to enhance LLM performance~\cite{hanparameter}.

\subsection{Limitations of Next-Token Prediction}
\label{sec:limitations_of_next_token_prediction}
However, task performance is not solely determined by parametric knowledge $K$. The expression mechanism, $E_{\textit{ntp}}$, is a vital component of the task completion process. The NTP paradigm can constrain the use of $K$~\cite{gekhman2025inside}. Even when internal knowledge $K$ is sufficient for a task, the LLM is forced to manifest its capacity by generating textual outputs based on a probability distribution over the vocabulary. Consequently, the expressive power of $K$ is bottlenecked by the NTP expression mechanism ($E_{\textit{ntp}}$).

To illustrate this expression bottleneck, we examine three representative examples of tasks where $E_{\textit{ntp}}$ is sub-optimal:

\textbf{1. Information Retrieval} ($T_{\textit{IR}}$). 
Standard NTP outputs lack the dense, high-dimensional information required for semantic proximity matching, and the generation process is time-consuming. An expression mechanism that encodes inputs directly into high-quality embeddings is significantly more effective than $E_{\textit{ntp}}$.
This example represents scenarios where the required output cannot be adequately represented by sequential tokens.
    
\textbf{2. Object Detection} ($T_{\textit{OD}}$). 
Expressing coordinates of an object as tokenized strings introduces sequential dependency and potential syntax errors. Post-processing is necessary to make outputs usable. A more efficient expression is the direct regression of numerical values from the model’s latent representations.
This example represents scenarios where $E_{\textit{ntp}}$ may be functionally capable, but with low efficiency and cumbersome output formats.

\textbf{3. Anomaly Detection} ($T_{\textit{AD}})$. 
LLMs typically perform classification tasks like anomaly detection by generating label tokens (e.g., \emph{yes} or \emph{no}). However, the resulting token probabilities are often poor proxies for task-oriented confidence scores. Because the model operates over a vast vocabulary, probability mass frequently leaks to semantically similar tokens. An expression that uses task-oriented classification logits, provides explicit, normalized class probabilities, that are more robust than the scattered distribution of $E_{ntp}$.
This task represents scenarios where $E_{\textit{ntp}}$ fails to provide the necessary information for rigorous task evaluation.

These $E_{\textit{ntp}}$ constraints are critical, as such tasks are foundational to practical, domain-specific applications, including retrieval-augmented generation (RAG)~\cite{wei2024uniir,li2025fingear}, table localization~\cite{sui2024table}, and abnormal figure detection~\cite{xu2025towards,deng2025vmad}.

\subsection{Definition of Self Knowledge Re-expression}
\label{sec:task_definition_and_benefits_of_knowledge_re-expression}
Adapting LLMs to specialized tasks typically requires extensive, high-quality annotated data $D =  \langle D_x, D_y \rangle$, whether derived from human~\cite{chen2021finqa} or model-assisted~\cite{wu2025boosting} sources. 
However, we hypothesize that the LLM already possesses sufficient intrinsic knowledge $K$ to address these tasks, and that the performance gap is a failure of expression rather than a lack of knowledge. 

\label{sec:methods}
\begin{figure*}[t]
\centering
\begin{center}
   \includegraphics[width=0.98\linewidth]{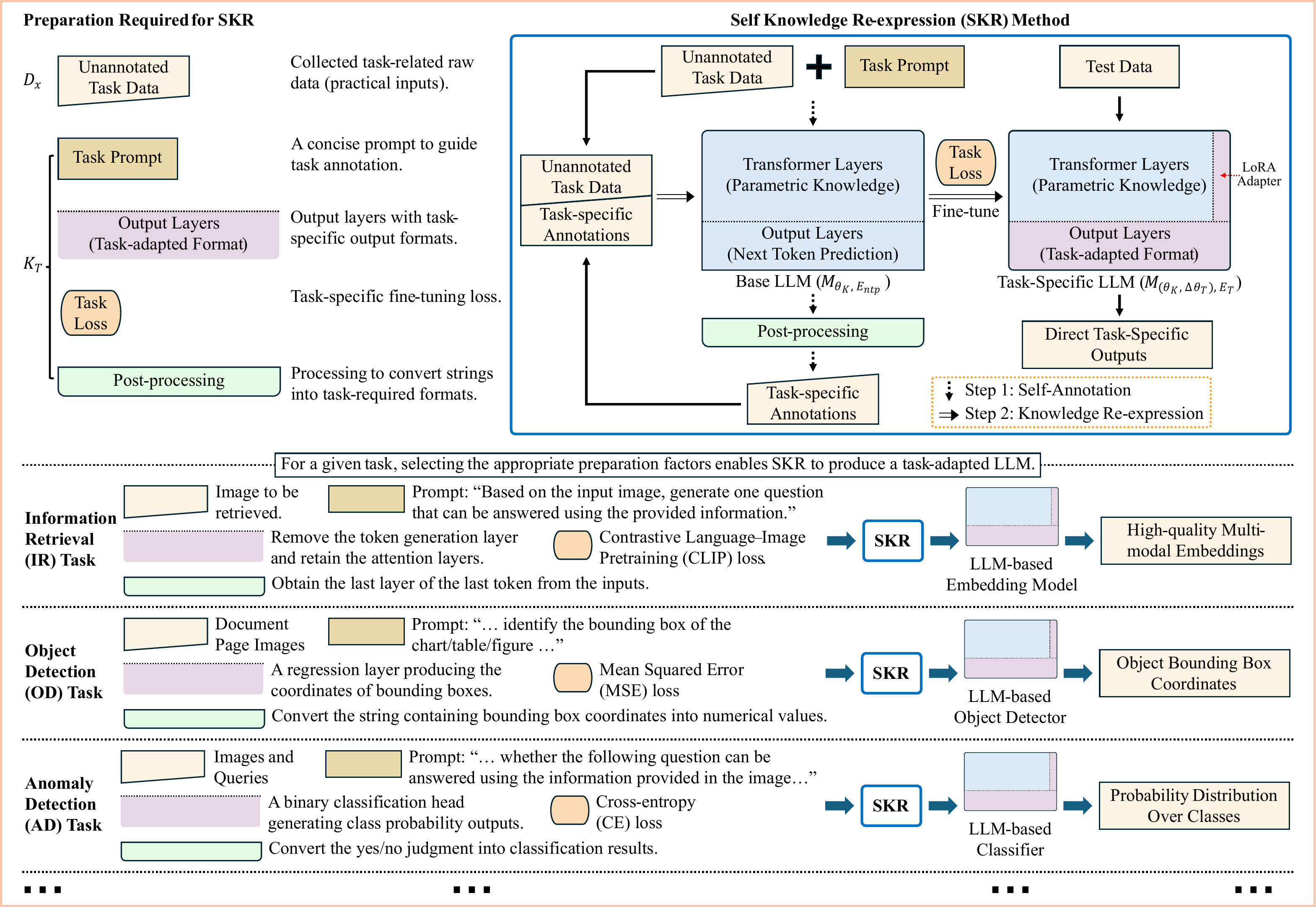}
    \vspace{-0.3cm}
   \caption{The Self Knowledge Re-expression (SKR) method implemented across three different tasks (i.e., information retrieval, objection detection, anomaly detection). The process executes using only unannotated raw data and task-related configurations (e.g., prompts, task loss, output formats, lightweight post-processing), without requiring human annotation or external supervision.}
\label{fig:method}
\end{center}
\vspace{-0.6cm}
\end{figure*}

We therefore formalize the \textbf{S}elf \textbf{K}nowledge \textbf{R}e-expression (\textbf{SKR}) process as follows. Given a task $T$ and only unannotated data $D_x$, SKR transforms LLM $M$'s expression from the generic $E_{\textit{ntp}}$ to a task-adapted mechanism $E_{T}$. This process relies primarily on $M$'s intrinsic knowledge $K$, without external supervision from humans or other LLMs. 
For rigor, we introduce $K_T$ to represent the task priors required for adaptation, such as expected output layer structure, loss function, and task prompts. While $K_T$ may include knowledge external to $K$, its informational content is negligible in scale ($K_T \ll K$). Unlike the extensive effort required for annotating $\mathcal{D}_x$, defining $K_T$ consists of only lightweight, high-level design choices. Therefore, $K_T$ does not undermine the premise that SKR primarily relies on the $M$'s intrinsic knowledge. Let $M_{E_*}$ denote the model $M$ using expression mechanism $E_*$. SKR is formalized as:
\begin{equation}
M_{E_{\textit{ntp}}} \underset{K, K_T, \mathcal{D}_x}{\xlongrightarrow{\text{\quad SKR \quad}}} M_{E_{T}}.
\end{equation}

SKR differs from established techniques like knowledge distillation or teacher-student paradigm in two aspects: (1) SKR relies on the LLM $M$'s intrinsic knowledge to achieve task adaptation, without the support of a superior ``teacher model'' to inject external knowledge. (2) SKR aims at optimizing the expression mechanism ($E_{\textit{ntp}} \to E_T$) to unlock latent potential, rather than introducing new knowledge to improve $K$.
These characteristics make SKR a highly economical method by significantly reducing adaptation costs. Furthermore, as a fully local process, it mitigates the privacy risks of external data exposure, making it particularly suitable for sensitive domains such as finance and healthcare.

\section{Methods}

To implement SKR, we propose a method enabling pre-trained LLMs to adapt to specialized tasks by optimizing their knowledge expression mechanism. Our method uses the LLM itself to generate task-specific annotations, guided by task-specific priors ($K_T$), to transition from generic token generation to task-optimized outputs (Figure~\ref{fig:method}).


\subsection{SKR Method}


The SKR method executes in two sequential steps:

\textbf{1. Self-Annotation: Knowledge Extraction via $E_{\textit{ntp}}$}. The process begins with unannotated task data $\mathcal{D}_x$. The original LLM is queried to generate target outputs in a sequential token format (e.g., image-related text for $T_{\textit{IR}}$, or coordinate strings for $T_{\textit{OD}}$). Here, $E_{\textit{ntp}}$ serves not as the final output method, but as a mechanism for knowledge extraction. Post-processing converts outputs into the task-specific format, producing self-generated annotations $\mathcal{\hat{D}}_y$ for $\mathcal{D}_x$:
\begin{equation}
\label{eq:self_annotation} (\mathcal{D}_x, T) \underset{\text{Task Query}}{\xlongrightarrow{M_{E_{\textit{ntp}}}}} \mathcal{\hat{D}}_y.
\end{equation}

\textbf{2. Knowledge Re-expression: Transitioning to $E_T$}. Using the self-annotated data $(\mathcal{D}_x, \mathcal{\hat{D}}_y)$, the model is fine-tuned with a task-specific objective (e.g., contrastive loss for embeddings, or mean squared error loss for regression). This step transforms the model's output mechanism from $E_{\textit{ntp}}$ to a task-suitable expression $E_T$. We implement this process using LoRA~\cite{hu2022lora}. With original parameters $\theta_K$ and trainable parameters $\Delta \theta_T$ (comprising LoRA adapters and a task-specific head), the adaptation is formalized as:
\begin{equation}
M_{\theta, E_{\textit{ntp}}} \underset{(\mathcal{D}_x, \mathcal{\hat{D}}_y), \text{ Loss of }E_T}{\xlongrightarrow{\text{\quad\quad Fine-tune \quad\quad}}} M_{(\theta, \Delta \theta_T), E_T}.
\end{equation}
This step results in a task-adapted model $M_{(\theta, \Delta \theta_T), E_T}$.

\subsection{Case Studies and Task Implementations}
We implement our SKR method on the three representative task types (Figure~\ref{fig:method}). While LLMs and multi-modal LLMs (MLLMs) share the same expression mechanism, we prioritize MLLMs experiments due to their broader application potential. Our evaluation covers both single- and multi-modal scenarios: $T_{\textit{IR}}$ includes a text-only variant to validate SKR on pure LLMs. All three task types are implemented with multi-modal inputs to demonstrate MLLM adaptation. 

\subsubsection{SKR for Extended Capability ($T_{\textit{IR}}$)}
The first case study uses the IR tasks to highlight capabilities that standard NTP does not support. We cover both text-to-text (paragraph) and text-to-image (chart/figure) retrieval, where the objective is to identify elements containing the information to answer a user query. The generic $E_{\textit{ntp}}$ expression lacks the ability to produce vectors required for efficient semantic search, especially in multi-modal contexts~\cite{jiang2024e5}. Standard solutions rely on fine-tuning using text-answer pairs created by humans or superior teacher models~\cite{zhou2024vista,koukounas2024jina}.

Our SKR method only needs raw paragraphs or images. We prompt the model to generate queries based on these inputs. The queries and inputs then form the training pairs for contrastive learning~\cite{radford2021learning}. In the Knowledge Re-expression phase, we bypass NTP layers and use the final token's last layer as the representation. Fine-tuning with a contrastive loss transitions the model from $M_{\theta, E_{\textit{ntp}}}$ to $M_{(\theta, \Delta \theta_{\textit{IR}}), E_{\textit{IR}}}$.
This transition resolves the $E_{\textit{ntp}}$ bottleneck, making SKR-adapted models surpass advanced baselines.

\subsubsection{SKR for Efficiency ($T_{\textit{OD}}$)}
The study of $T_{\textit{OD}}$ demonstrates how SKR streamlines multi-step tasks into efficient end-to-end operations. Practical object detection often requires flexible granularity. For example, extracting a chart requires including the title and source for a full context. 
Traditional detectors, like YOLO~\cite{redmon2016you}, lack zero-shot instruction flexibility. Although MLLMs can adjust detection objectives by adding instructions, they often struggle to detect such ``composite'' areas (figure + title + source). Current workarounds split detection into sub-tasks (e.g., separate title and figure detection) and merge results via post-processing, which dramatically increases latency and risk of parsing failures.

Through SKR, we implement \textit{Self-Annotation} by running multi-step detection on a small data subset to generate coordinates. We then replace the NTP layers with a regression head $E_{\textit{OD}}$ that directly outputs bounding box values. Fine-tuned with an MSE loss, the adapted model $M_{(\theta, \Delta \theta_{\textit{OD}}), E_{\textit{OD}}}$ enables end-to-end detection. This process eliminates multi-turn detection and post-processing, significantly reducing latency. Crucially, for models lacking inherent grounding abilities (e.g., LLaVA), SKR effectively evolves latent image understanding into explicit grounding capability.

\subsubsection{SKR for More Information ($T_{\textit{AD}}$)}
This case study transforms the expression mechanism to provide task-specific information. Our anomaly detection task is an image-text matching problem: determining if an image contains the information relevant to a specific query. Since valid matches are rare, the data is highly imbalanced. $E_{\textit{ntp}}$ restricts output to next-token probabilities, explicit probability distribution over target classes is missing.

In the \textit{Self-Annotation} phase, we prompt the model to output an explicit \emph{yes} or \emph{no} for query-image pairs. In the \textit{Knowledge Re-expression} phase, we replace the generation head with a binary classification head ($E_{\textit{AD}}$) that outputs normalized class probabilities. This transitions the model from NTP to providing class distributions, allowing for the calculation of robust metrics like AUROC and AUPRC. These metrics are important for imbalanced datasets~\cite{hu2020hrn,wang2022cmg} and impossible to derive accurately from simple token generation.


\section{Experimental Setting}
\label{sec:exp_settings}

\subsection{Dataset}
\label{sec:dataset}
We evaluate SKR using a large-scale industrial-scale financial dataset and the open-source MMDocRAG dataset~\cite{li2025closing}, DocVQA-2020 dataset~\cite{mathew2021docvqa}, SciMMIR dataset~\cite{wu2024scimmir}, and CUB\_200\_2011 dataset~\cite{wah2011caltech}. For all datasets, SKR uses only unannotated raw data. Annotations are strictly reserved for performance evaluation. More detailed descriptions are provided in Appendix~\ref{sec:dataset_details}.

\textbf{Financial Document Dataset}: This industrial-scale dataset comprises over 100,000 professional financial reports, including millions of analysis paragraphs, tables, and charts. These documents provide a rigorous environment for $T_{\textit{IR}}$, $T_{\textit{OD}}$, and $T_{\textit{AD}}$, reflecting the complexities of real-world financial analysis. While the total corpus is vast, the SKR method requires only a small, randomly sampled subset of raw data ($\approx$1,000 samples) to serve as the unannotated training set $\mathcal{D}_x$. To ensure robust evaluation, we constructed specialized test sets for each task, all subject to meticulous manual verification. For $T_{\textit{IR}}$, the model retrieves relevant images from a candidate pool in response to specific textual queries. For $T_{\textit{OD}}$, the model identifies and localizes charts and tables within a document page, outputting precise bounding boxes. For $T_{\textit{AD}}$, the model performs a binary verification task, determining whether a given query and image form a valid question-answer pair. 

\textbf{MMDocRAG Dataset}: We evaluate the multi-modal IR task on the open-source MMDocRAG dataset. We strictly follow their evaluation protocols but use only training data without labels. This allows for a direct comparison with state-of-the-art models in a standardized RAG environment. 

\textbf{DocVQA-2020 Dataset and SciMMIR Dataset}: To assess the generalization of SKR, we test models, which are fine-tuned only on MMDocRAG, directly on DocVQA-2020 and SciMMIR. DocVQA-2020 includes documents spanning multiple decades and diverse industries. SciMMIR is a specialized scientific domain dataset. This allows us to evaluate if SKR models can generalize to unseen datasets for similar tasks.

\textbf{CUB\_200\_2011 dataset}: For the OD and AD tasks, we further evaluate SKR on this open-source dataset. It contains 200 bird species with fine-grained bounding box annotations, providing a standard environment to assess SKR’s ability to localize objects and identify anomalies.


\subsection{Evaluation Metrics}
We evaluate each task using a comprehensive set of metrics, with definitions and equations detailed in Appendix~\ref{sec:evaluation_metrics}. For $T_{\textit{IR}}$, we report mean reciprocal rank (MRR) and Recall@k (R@k). 
For $T_{\textit{OD}}$, we report intersection over union (IoU) and inference time (Time). For $T_{\textit{AD}}$, we use accuracy (Acc), area under the ROC curve (AUROC), area under the precision-recall curve (AUPRC), and inference time (Time).

\subsection{Models and Baselines}

We evaluate four popular LLMs for the text-only task and three MLLMs for the multi-modal tasks. LLMs include Mistral-7B-Instruct-v0.2 (Mistral-0.2; \citealt{jiang2023clip}), Phi-3-mini-128k-instruct (Phi-3; \citealt{abdin2024phi}), Qwen2-7B-Instruct (Qwen2; \citealt{team2024qwen2}), and Llama-3.1-8B (Llama-3.1; \citealt{dubey2024llama}). MLLMs include LLaVA-v1.6-Mistral-7B (LLaVA-1.6; \citealt{liu2023visual}), Phi-3-vision-128k-instruct (Phi-3-VL; \citealt{abdin2024phi}), and Qwen2-VL-7B-Instruct (Qwen2-VL; \citealt{wang2024qwen2}). These models serve as the base models $M$, where the generic $E_{\textit{ntp}}$ is adapted to a task-specific $E_T$ via SKR.

Our primary baseline is the base model before SKR, which isolates the gains strictly attributable to our method. Since SKR operates in a zero-annotation regime, few established adaptation methods are directly comparable. We include E5-V~\cite{jiang2024e5} (a training-free representation method) and several state-of-the-art embedding models (Nomic~\cite{nomicembedmultimodal2025}, OpenAI, and top-tier models from the MMDocRAG benchmark) that are specifically fine-tuned on massive datasets for IR, to demonstrate SKR's benefits.
We also compare with models fine-tuned on external supervision (e.g., GPT-4o) and implement an SFT baseline using the same self-annotations to facilitate the ablation study.

Although our study does not use the latest open-source models due to resource constraints, our results already surpass many specialized systems, suggesting the possibility of further improvement on better LLMs. However, our primary goal is not to explore the upper bounds of SOTA performance, but to provide a controlled proof of the existence of the expression bottleneck. By comparing the same model under $E_{\text{ntp}}$ and $E_T$, we show that diverse LLMs possess vast unexploited capabilities that can be unlocked simply by optimizing their expression mechanisms, allowing even modest models to outperform highly specialized counterparts.

\subsection{SKR Implementation Details}


The \textit{Self-Annotation} phase is tailored to the base model's inherent capabilities.
For $T_{\textit{IR}}$ and $T_{\textit{AD}}$, we directly prompt the model to generate input-related queries or assess image-text relevance.
For $T_{\textit{OD}}$, the strategy diverges by model capability. We directly prompt Qwen2-VL model to output coordinates since it possesses native grounding capabilities. In contrast, models with semantic understanding but no native coordinate generation (LLaVA-1.6 and Phi-3-VL) use a ``fuzzy localization'' strategy (i.e., recursively cropping inputs and querying object presence) to get bounding boxes. This distinction allows us to prove that SKR can not only optimize existing capabilities but also awaken latent skills in models that lack native support. Additionally, we also experiment with using lightweight, rule-based OCR as a post-processing step to assist in coordinate refinement.

Further details on self-annotation, prompts, task head structures, and hyper-parameters are provided in Appendix~\ref{sec:more_implementation_details}.

\section{Results and Discussion}
\label{sec:results_discussions}

\begin{table}[t]
\vspace{-0.0cm}
\renewcommand\arraystretch{1}
\caption{Performance comparison on $T_{\textit{IR}}$. SKR's results are highlighted in bold. Across all models and modalities, SKR consistently achieves the best performance. Qwen2 models adapted via SKR outperform specialized industrial baselines.}
\vspace{-0.2cm}
\begin{center}
\small
\begin{tabular}[b]{l@{\hspace{3pt}} l r@{\hspace{10pt}} r@{\hspace{10pt}} r@{\hspace{10pt}} r@{\hspace{10pt}} r@{\hspace{10pt}}}
\toprule
\multicolumn{2}{l}{Model} & MRR & R@1 & R@3 & R@5 & R@10 \\
\cmidrule{1-7}
\multicolumn{7}{l}{Text-to-text Retrieval} \\
\cmidrule{1-7}
\multicolumn{2}{l}{OpenAI ada-002}   & 0.420 & 0.322 & 0.460 & 0.518 & 0.598 \\
\multicolumn{2}{l}{Nomic-text-v1.5}    & 0.495 & 0.392 & 0.558 & 0.624 & 0.702 \\
\multicolumn{2}{l}{Nomic-text-v2}  & 0.549 & 0.454 & 0.600 & 0.648 & 0.728 \\

\cmidrule{1-2}
\multirow{4}{*}{Mistral-0.2}
 & Base & 0.108 & 0.052 & 0.116 & 0.154 & 0.204 \\
 & E5-V & 0.223 & 0.144 & 0.250 & 0.292 & 0.386 \\
 & SFT  & 0.076 & 0.036 & 0.078 & 0.104 & 0.152 \\
 & \textbf{SKR}  & \textbf{0.622} & \textbf{0.510} & \textbf{0.702} & \textbf{0.754} & \textbf{0.822} \\
\cmidrule{1-2}
\multirow{4}{*}{Phi-3}
 & Base       & 0.098 & 0.068 & 0.096 & 0.112 & 0.146 \\
 & E5-V       & 0.209 & 0.142 & 0.230 & 0.270 & 0.338 \\
 & SFT  & 0.082 & 0.054 & 0.084 & 0.100 & 0.124 
 \\
 & \textbf{SKR}        & \textbf{0.632} & \textbf{0.518} & \textbf{0.718} & \textbf{0.778} & \textbf{0.826} \\
\cmidrule{1-2}
\multirow{4}{*}{Qwen2}
 & Base       & 0.115 & 0.062 & 0.130 & 0.164 & 0.230 \\
 & E5-V       & 0.270 & 0.196 & 0.294 & 0.336 & 0.416 \\
 & SFT        & 0.063 & 0.032 & 0.056 & 0.090 & 0.136 \\
 & \textbf{SKR}        & \textbf{0.674} & \textbf{0.566} & \textbf{0.756} & \textbf{0.814} & \textbf{0.860} \\
\cmidrule{1-2}
\multirow{4}{*}{Llama3.1}
 & Base    & 0.196 & 0.136 & 0.214 & 0.258 & 0.330 \\
 & E5-V    & 0.306 & 0.240 & 0.318 & 0.360 & 0.426 \\
 & SFT     & 0.197 & 0.140 & 0.210 & 0.256 & 0.328 \\
 & \textbf{SKR}     & \textbf{0.669} & \textbf{0.566} & \textbf{0.742} & \textbf{0.792} & \textbf{0.848} \\

\cmidrule{1-7}
\multicolumn{7}{l}{Text-to-image Retrieval} \\
\cmidrule{1-7}
\multicolumn{2}{l}{Nomic-mm-7b}  & 0.669 & 0.560 & 0.745 & 0.795 & 0.860 \\

\cmidrule{1-2}
\multirow{4}{*}{LLaVA-1.6}
 & Base & 0.014 & 0.000 & 0.010 & 0.030 & 0.035 \\
 & E5-V & 0.197 & 0.150 & 0.205 & 0.230 & 0.305 \\
 & SFT  & 0.031 & 0.020 & 0.025 & 0.030 & 0.065 \\
 & \textbf{SKR}  & \textbf{0.552} & \textbf{0.445} & \textbf{0.625} & \textbf{0.665} & \textbf{0.730} \\
\cmidrule{1-2}
\multirow{4}{*}{Phi-3-VL}
 & Base & 0.071 & 0.050 & 0.065 & 0.080 & 0.125 \\
 & E5-V & 0.203 & 0.150 & 0.210 & 0.235 & 0.300 \\
 & SFT  & 0.027 & 0.020 & 0.030 & 0.030 & 0.035 \\
 & \textbf{SKR}  & \textbf{0.556} & \textbf{0.450} & \textbf{0.630} & \textbf{0.680} & \textbf{0.750} \\
\cmidrule{1-2}
\multirow{4}{*}{Qwen2-VL}
 & Base & 0.044 & 0.030 & 0.035 & 0.040 & 0.090 \\
 & E5-V & 0.291 & 0.210 & 0.340 & 0.395 & 0.450 \\
 & SFT  & 0.055 & 0.035 & 0.055 & 0.065 & 0.085 \\
 & \textbf{SKR}  & \textbf{0.691} & \textbf{0.595} & \textbf{0.755} & \textbf{0.795} & \textbf{0.875} \\

\bottomrule
\end{tabular}
\end{center}
\label{tab:ir_results}
\vspace{-0.9cm}
\end{table}

We analyze the performance gains of SKR-adapted models relative to their original base versions and established industrial baselines across $T_{\textit{IR}}$, $T_{\textit{OD}}$, and $T_{\textit{AD}}$ in Section~\ref{sec:task_improvement_of_skr}. The comparative analysis between SKR and standard SFT is presented in Section~\ref{sec:ablation_study} to demonstrate the necessity of transitioning from NTP to a task-specific expression head. 

\subsection{Task Improvement of SKR}
\label{sec:task_improvement_of_skr}

\subsubsection{SKR for Extended Capability ($T_{\textit{IR}}$)}
\label{sec:skr_tir}
Table~\ref{tab:ir_results} demonstrates that SKR yields a transformative improvement in the semantic representation capabilities of both LLMs and MLLMs. Under the standard $E_{\textit{ntp}}$ paradigm, base models perform poorly on retrieval tasks because they are not optimized to generate high-quality representations. While the prompt-based method E5-V mitigates this issue to an extent, its performance remains insufficient. 

Our SKR method directly adapts base models into high-quality encoding models. Without any external annotations, SKR-adapted Qwen2 models consistently outperform industrial models trained on massive annotated datasets.
More evaluations on open-source benchmarks further validate these findings (Appendix~\ref{sec:ir_opensource_results}). Results on the open-source MMDocRAG dataset demonstrate that SKR consistently outperforms various SOTA retrieval models. Results on the DocVQA and SciMMIR datasets underscore the robust cross-dataset generalization ability of SKR-adapted models across similar tasks.
All these results proves that the bottleneck in current retrieval systems is often not a scarcity of knowledge, but rather the inability of the expression mechanism to properly use knowledge. 

\begin{table}[t]
\vspace{-0.0cm}
\renewcommand\arraystretch{1}
\caption{Performance comparison on $T_{\textit{OD}}$. Results for the SKR method are highlighted in bold. The best results within each model are underlined. Runtime is reported in minutes per 100 test cases.}
\begin{center}
\small
\begin{tabular}[b]{l@{\hspace{3pt}} l r r}
\toprule
\multicolumn{2}{l}{Model} & IoU & Time \\
\cmidrule{1-4}

\multirow{5}{*}{LLaVA-1.6}
 & Base (Fuzzy Location) & 0.142 & \quad 41.0  \\
 & SFT (Fuzzy Location) & 0.065 & \quad 6.4  \\
 & \textbf{SKR} (Fuzzy Location) & \textbf{0.266} & \quad \textbf{1.7}  \\
 & Base (OCR) & 0.599 & \quad 10.3  \\
 & \textbf{SKR} (OCR) & \underline{\textbf{0.614}} & \quad \underline{\textbf{1.6}}  \\

\cmidrule{1-2}
\multirow{5}{*}{Phi-3-VL}
 & Base (Fuzzy Location) & 0.164 & \quad 35.0  \\
 & SFT (Fuzzy Location) & 0.056 & \quad 7.1  \\
 & \textbf{SKR} (Fuzzy Location) & \textbf{0.232} & \quad \underline{\textbf{0.8}}  \\
 & Base (OCR) & 0.336 & \quad 30.7  \\
 & \textbf{SKR} (OCR) & \textbf{\underline{0.344}} & \quad \underline{\textbf{0.9}}  \\

\cmidrule{1-2}
\multirow{4}{*}{Qwen2-VL}
 & Base (One-turn Prompt) & 0.628 & \quad 6.8  \\
 & Base (Multi-turn Prompt) & 0.739 & \quad 20.4  \\
 & SFT (Multi-turn Prompt) & \underline{0.741} & \quad 6.5  \\
 & \textbf{SKR} (Multi-turn Prompt) & \textbf{0.726} & \quad \underline{\textbf{4.8}}  \\

\bottomrule
\end{tabular}
\end{center}
\label{tab:od_results}
\vspace{-0.6cm}
\end{table}

\subsubsection{SKR for Efficiency ($T_{\textit{OD}}$)}
\label{sec:skr_tod}

The results in Table~\ref{tab:od_results} highlight SKR’s ability to elicit latent capabilities and streamline multi-step pipelines. Models such as LLaVA-1.6 and Phi-3-VL, which cannot localize objects, acquire object detection capability through SKR. Basic fuzzy location post-processing enables models to output more reasonable coordinates. Adding simple OCR yields substantial improvements. Both models exhibit dramatic IoU improvements of 0.472 and 0.180, respectively, compared to their base states.

For Qwen2-VL, which inherently possesses grounding capabilities, SKR compresses multi-turn actions into a single turn. The adapted model achieves an IoU of 0.726, significantly outperforming the original model’s one-turn result and remaining comparable to the multi-turn performance. Across all models, SKR reduces inference time by at least 76\%. Evaluations on the open-source CUB\_200\_2011 dataset demonstrate a consistent trend (see Appendix~\ref{sec:od_ad_opensource_results}). Replacing $E_{\textit{ntp}}$ with $E_{\textit{reg}}$ eliminates the computational overhead of sequential token generation.

\subsubsection{SKR for More Information ($T_{\textit{AD}}$)}
\label{sec:skr_tad}
In anomaly detection, where datasets are highly imbalanced, standard accuracy metrics can be deceptive. Table~\ref{tab:ad_results} reveals that SKR provides a significantly more robust output mechanism than binary token prediction. 
To ensure a rigorous baseline, we tried two probability extraction methods for the base models: assigning 100\% probabilities to generated classes and getting softmax probabilities for specific tokens. We report the superior of these two as base model results.

While accuracy gains appear modest due to the rarity of outliers, AUROC and AUPRC scores, which are more informative for anomaly detection, show massive gains. All models achieve at least a 0.1 improvement in AUROC and a 0.3 increase in AUPRC; Phi-3-VL’s AUPRC jumps from 0.287 to 0.933. The performance benefit is further corroborated by evaluations on the open-source CUB\_200\_2011 benchmark (see Appendix~\ref{sec:od_ad_opensource_results}). This confirms that $E_{\textit{ntp}}$ is ill-suited for providing the calibrated confidence scores necessary for such tasks. Additionally, SKR-adapted models demonstrate enhanced efficiency, reducing runtime by at least 30\%.

\begin{table}[t]
\vspace{-0.2cm}
\renewcommand\arraystretch{1}
\caption{Performance comparison on $T_{\textit{AD}}$. SKR results are highlighted in bold. SKR-adapted models consistently achieve the highest performance with lowest runtime. Runtime is reported in minutes per 100 test cases.}
\vspace{-0.0cm}
\begin{center}
\small
\begin{tabular}[b]{l@{\hspace{3pt}} l r@{\hspace{10pt}} r@{\hspace{10pt}} r@{\hspace{10pt}} r@{\hspace{10pt}} r@{\hspace{10pt}}}
\toprule
\multicolumn{2}{l}{Model} & Acc & AUROC & AUPRC & Time \\
\cmidrule{1-6}
\multirow{3}{*}{LLaVA-1.6}
 & Base & 0.930 & 0.894 & 0.531 & 3.8 \\
 & SFT & 0.835 & 0.892 & 0.381 & 1.4 \\
 & \textbf{SKR}  & \textbf{0.975} & \textbf{0.991} & \textbf{0.879} & \textbf{1.3} \\
\cmidrule{1-2}
\multirow{3}{*}{Phi-3-VL}
 & Base & 0.770 & 0.850 & 0.287 & 4.7 \\
 & SFT & 0.945 & 0.912 & 0.850 & 7.1 \\
 & \textbf{SKR}  & \textbf{0.935} & \textbf{0.981} & \textbf{0.933} & \textbf{0.7} \\
\cmidrule{1-2}
\multirow{3}{*}{Qwen2-VL}
 & Base & 0.935 & 0.875 & 0.532 & 1.1 \\
 & SFT & 0.960 & 0.939 & 0.746 & 0.9 \\
 & \textbf{SKR}  & \textbf{0.960} & \textbf{0.984} & \textbf{0.868} & \textbf{0.7} \\

\bottomrule
\end{tabular}
\end{center}
\label{tab:ad_results}
\vspace{-0.8cm}
\end{table}

\subsection{Ablation Studies}
\label{sec:ablation_study}
To investigate the relationship between an LLM's latent knowledge and its functional capabilities, we conduct ablation studies on the two SKR steps. Results reveals that an LLM's internal knowledge sufficiency is contingent upon: (i) model's intrinsic knowledge depth, (ii) model's pre-existing task-specific abilities, and (iii) task complexity.

\subsubsection{Impact of Annotation Knowledge: Intrinsic vs. External}
\label{sec:comparison_with_external}
\begin{figure}[t]
\centering
\vspace{-0.0cm}
\begin{center}
   \includegraphics[width=0.98\linewidth]{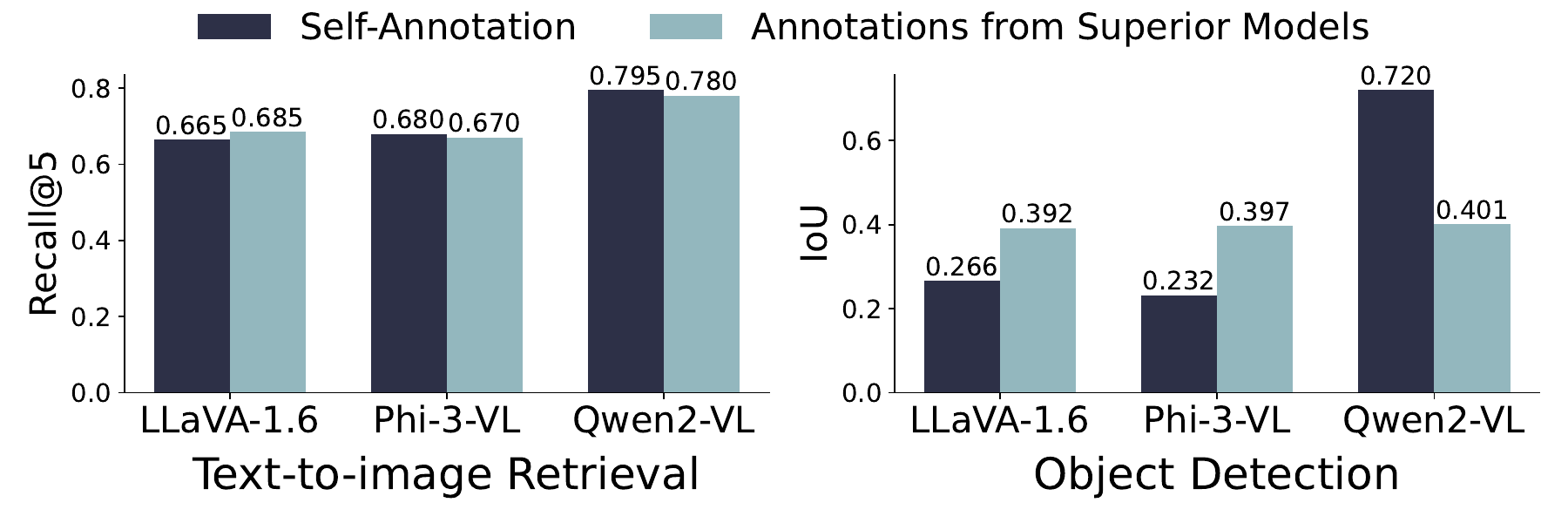}
    \vspace{-0.4cm}
   \caption{Performance comparison of LLMs fine-tuned using self-annotations versus external annotations across $T_{\textit{IR}}$ and $T_{\textit{OD}}$.}
\label{fig:knowledge_source_comparison}
\end{center}
\vspace{-0.8cm}
\end{figure}

We investigate the sufficiency of intrinsic knowledge by comparing it with external supervision. We use GPT-4o, which is stronger than our evaluated models on multi-modal benchmarks~\cite{zhu2025multichartqa}, to generate annotations for $T_{\textit{IR}}$ and $T_{\textit{OD}}$. We use the same re-expression architectures ($E_T$) of SKR, changing only annotations.
As illustrated in Figure~\ref{fig:knowledge_source_comparison}, this comparison clarifies where internal knowledge suffices and where external knowledge remains beneficial. 

For $T_{\textit{IR}}$, self-annotation yields performance gains comparable to, and occasionally surpassing, external supervision. Since image understanding is a foundational capability of modern MLLMs, the gap between our evaluated models and GPT-4o in generating image-related text is marginal. This confirms that the retrieval bottleneck stems primarily from the expression mechanism rather than knowledge deficiency.

In contrast, $T_{\textit{OD}}$ results vary by model. External supervision works better for LLaVA-1.6 and Phi-3-VL, as these models lack native grounding capabilities. However, Qwen2-VL’s self-annotation outperforms GPT-4o’s labels. Despite GPT-4o’s general dominance across benchmarks, its grounding ability (IoU of 0.414 on our test set) is inferior to Qwen2-VL’s specialized capability.
Therefore, while GPT-4o provides useful knowledge for models with clear gaps, it cannot match the performance of Qwen2-VL when SKR re-expresses Qwen2-VL's grounding knowledge.
This observation is reinforced by our analysis of the Gemma-3 family (Appendix~\ref{sec:model_size_analysis}), which indicates that larger models do not always possess more task potential than smaller models.

In summary, external supervision is redundant in many scenarios, where poor task performance stems from a failure of expression rather than a lack of knowledge. However, SKR is not a total substitute for knowledge acquisition. For models with clear knowledge gaps, external supervision remains essential. Since pre-training cannot encompass all tasks, these findings suggest that even leading LLMs likely possess untapped abilities, waiting to be correctly expressed.

\subsubsection{Impact of Expression Mechanism: \quad\quad\quad\quad NTP vs. Task-specific Head}
\label{sec:comparison_with_sft}
To isolate the impact of the expression mechanism, we compare SKR’s re-expression step against standard SFT. The SFT baseline uses the same self-annotations as SKR but fine-tunes the models via their native NTP objective. Results are summarized in the SFT rows of Tables~\ref{tab:ir_results}, \ref{tab:od_results}, \ref{tab:ad_results}.

In $T_{\textit{IR}}$, SFT yields poor results, with marginal gains in text-to-image retrieval and degradation in text-to-text retrieval. These results underscore a fundamental objective mismatch: SFT forces the model to express knowledge through sequential token generation, whereas $T_{\textit{IR}}$ requires high-quality representation vectors. SKR’s task-specific head bypasses this bottleneck by directly using latent representations.

The effectiveness of SFT in $T_{\textit{OD}}$ depends heavily on pre-existing abilities. While SFT improves Qwen2-VL, it severely degrades LLaVA-1.6 and Phi-3-VL. Lacking native grounding capabilities, these two models struggle to produce structured coordinate strings via NTP without the constraints of a task-specific head. 
Furthermore, even for Qwen2-VL, SFT remains 35\% slower than SKR. Although SFT also compressed multi-turn detection into a single-turn running, the NTP mechanism still requires multiple decoding steps for token generation, whereas SKR produces the entire bounding box in a single forward pass.

For $T_{\textit{AD}}$, performance remains substantially inferior to SKR. In terms of AUPRC, the most critical metric for imbalanced classification, SKR consistently outperforms SFT by at least 8\%, proving that direct class projections are more effective than mapping classifications to vocabulary tokens.

These findings demonstrate that a dedicated re-expression mechanism is vital for model adaptation. SKR’s ability to extract latent knowledge and optimize inference efficiency provides a decisive advantage that standard NTP-based fine-tuning cannot replicate. Further investigations into the necessity of LoRA adapters for facilitating the re-expression process, as well as an analysis of how SKR reshapes latent representations, are provided in Appendix~\ref{sec:lora_ablation} and Appendix~\ref{sec:cka_analysis}, respectively.

\section{Conclusions and Future Work}
\label{sec:conclusions}
This paper formalizes the knowledge expression bottleneck, demonstrating that NTP often constrains LLMs' latent potential. We propose the SKR method, which is task-agnostic and fully local, to elicit this potential by transitioning from generic token generation to task-specific expressions. SKR significantly reduces adaptation costs and mitigates data exposure risks. Experiments across three task types show that SKR achieves substantial performance and latency gains without external supervision, even surpassing leading industrial models. Ablation studies confirm that external supervision is often redundant when the necessary knowledge resides within the model's parameters, proving that the expression bottleneck is a primary factor limiting LLM utility.

By decoupling knowledge from linguistic outputs, this work establishes knowledge re-expression as a novel adaptation paradigm. This shift opens promising research directions, such as the full automation of SKR, where models autonomously design their outputs. Furthermore, our findings suggest a path for evolving advanced LLMs without continuous scaling of data or parameters. These insights offer a practical, efficient, and secure route for deploying state-of-the-art models in sensitive, real-world applications.


\section*{Impact Statement}
This paper presents work whose goal is to advance the field of 
machine learning. There are many potential societal consequences 
of our work, none which we feel must be specifically highlighted here.

\section*{Disclaimer}
This paper was prepared for informational purposes in part by the Machine Learning Center of Excellence group of JPMorgan Chase \& Co. and its affiliates ("JP Morgan”) and is not a product of the Research Department of JP Morgan. JP Morgan makes no representation and warranty whatsoever and disclaims all liability, for the completeness, accuracy or reliability of the information contained herein. This document is not intended as investment research or investment advice, or a recommendation, offer or solicitation for the purchase or sale of any security, financial instrument, financial product or service, or to be used in any way for evaluating the merits of participating in any transaction, and shall not constitute a solicitation under any jurisdiction or to any person, if such solicitation under such jurisdiction or to such person would be unlawful.

\bibliography{ref}
\bibliographystyle{icml2025}

\newpage
\appendix

\textbf{Appendix Overview} 
This Appendix provides comprehensive supplementary material to support the methodology, experimental setup, and extended analysis.
We first present technical details in Appendix~\ref{sec:dataset_details}, \ref{sec:evaluation_metrics}, and~\ref{sec:more_implementation_details}, which cover dataset specifications, formal definitions, and implementation details of the evaluation metrics, respectively.
Subsequently, we present extensive additional experimental results to offer deeper insights into the SKR method. These include:
\begin{itemize}
\item Performance evaluations across the three core tasks using the open-source MMDocRAG, DocVQA-2020, SciMMIR, and CUB\_200\_2011 datasets (Appendix~\ref{sec:open_source_data_results}).
\item An investigation into the impact of model scaling on SKR performance (Appendix~\ref{sec:model_size_analysis}).
\item Additional ablation studies regarding the necessity of LoRA adapters (Appendix~\ref{sec:lora_ablation}).
\item A comparative study of latent representation shifts using CKA analysis (Appendix~\ref{sec:cka_analysis}).
\end{itemize}


\section{Dataset Details}
\label{sec:dataset_details}

We evaluate the SKR method using two distinct data sources: a large-scale proprietary document dataset and the open-source MMDocRAG benchmark. This dual evaluation demonstrates SKR's effectiveness in both professional, practical applications and standardized academic benchmarks. 

\subsection{Financial Document Dataset}
All three task types ($T_{\textit{IR}}$, $T_{\textit{OD}}$, $T_{\textit{AD}}$) are implemented using our proprietary financial document dataset. As highlighted in Section~\ref{sec:dataset}, we use only a small, randomly sampled subset of raw, unannotated data to activate the SKR process. To ensure rigorous evaluation, all ground-truth labels in the test sets were established through the consensus of at least two human experts.

\textbf{Information Retrieval ($T_{\textit{IR}}$):} For text-to-text retrieval, we sample 1,000 paragraphs as unannotated training data for self-annotation. For testing, we construct a candidate pool of 5,000 distinct paragraphs, within which we identify 500 ground-truth query-paragraph pairs verified by humans. For text-to-image retrieval, we sample 1,100 images (including charts, tables, and figures) as the unannotated task data. The test set consists of a candidate pool of 5,000 images, with 200 human-verified query-image pairs.

\textbf{Object Detection ($T_{\textit{OD}}$):} We sample 800 document pages, each containing a chart, table, or figure, as the unannotated task data $\mathcal{D}_x$. For the test set, we separately sample 200 pages and manually annotate the bounding boxes of the target objects.

\textbf{Anomaly Detection ($T_{\textit{AD}}$):} We reuse the 1,100-image set from $T_{\textit{IR}}$ as the unannotated training data. The test set is derived from the $T_{\textit{IR}}$ test set, constructed to simulate a class-imbalanced anomaly detection scenario. It comprises 200 samples: 20 positive cases (images matching the query) and 180 negative cases (images irrelevant to the query), all manually verified.

\subsection{MMDocRAG}
MMDocRAG is one of the latest multi-modal benchmark designed to evaluate the retrieval and reasoning capabilities of MLLMs~\cite{dong2025benchmarking}. It encompasses 14,826 images across 222 documents and includes 4,055 expert-annotated question-answer pairs.
The original benchmark evaluates query-image retrieval on several state-of-the-art retrievers, including DSE models~\cite{ma2024unifying} and ColPali models~\cite{faysse2025colpali}. These baselines represent the current industrial standard for retrieval-augmented generation (RAG), as they are LLMs specifically fine-tuned on massive multi-modal datasets.

To validate SKR's competitiveness, we adhered strictly to the official MMDocRAG test protocols. However, our training regime is significantly more rigorous: SKR starts from base LLMs and uses only the raw images from the development set to fine-tune models. We deliberately ignore all expert-annotated questions to maintain a strictly annotation-free adaptation process. As shown in Appendix~\ref{sec:ir_opensource_results}, SKR-adapted models outperform all advanced models evaluated in the original benchmark.

\subsection{DocVQA-2020 and SciMMIR}
DocVQA-2020~\cite{mathew2021docvqa} includes documents spanning multiple decades and diverse industries. It consists of 50,000 questions defined on 12,000+ document images. SciMMIR~\cite{wu2024scimmir} is a specialized scientific domain dataset that presents a significant retrieval challenge, requiring the model to identify one relevant figure from a large candidate pool of 16,263 items.

To evaluate cross-dataset generalization, we use only the test sets of these two benchmarks. The evaluation is conducted using the SKR model exclusively fine-tuned on the unannotated training split of MMDocRAG, without any exposure to DocVQA or SciMMIR data during training. As detailed in the experimental results (see Appendix~\ref{sec:ir_opensource_results}), SKR models exhibit robust generalization, significantly outperforming advanced baseline RAG models.

\subsection{CUB\_200\_2011}
The CUB\_200\_2011 dataset~\cite{wah2011caltech} is a fine-grained image recognition benchmark containing 200 bird species, with a total of 5,994 training and 5,794 test samples. Each sample includes an image and its corresponding bounding box annotation. To format the AD task, we selected hummingbirds (4 species, covering 2\% of the data) as the anomalous class.

Consistent with our core methodology, we strictly adhere to an annotation-free regime during the training phase, using only the raw images from the training set without accessing any ground-truth labels or bounding boxes. Detailed performance results on this benchmark are presented in Appendix~\ref{sec:od_ad_opensource_results}.

\section{Evaluation Metrics}
\label{sec:evaluation_metrics}
This section provides the formal definitions and calculation formulas for the metrics used to evaluate the performance of SKR across the three task types.

\subsection{Information Retrieval ($T_{\textit{IR}}$)}
For retrieval tasks, we focus on the model's ability to rank the ground-truth item as high as possible within a candidate pool. Inference latency is not reported for these tasks, as encoding times remain comparable between base and SKR-adapted models.
Let $Q$ be the set of test queries, and for each query $q \in Q$, let $rank_q$ denote the position of the most relevant document/image in the retrieved list. 

\textbf{1. Mean Reciprocal Rank (MRR):} MRR calculates the average of the reciprocal ranks of the most relevant document for all queries: \begin{equation} \text{MRR} = \frac{1}{|Q|} \sum_{i=1}^{|Q|} \frac{1}{rank_i} \end{equation}

\textbf{2. Recall@k (R@k):} This metric measures the proportion of queries for which at least one relevant document/image is found within the top $k$ retrieved results.Given our assumption that a single "golden truth" exists for each query in our dataset, this metric reflects the probability that the ground-truth item is captured within the top $k$ results:
\begin{equation}
\text{Recall@k} = \frac{1}{|Q|} \sum_{i=1}^{|Q|} \mathbb{I}(rank_i \leq k)
\end{equation}
where $\mathbb{I}(\cdot)$ is the indicator function. We report results using $k \in \{1, 3, 5, 10\}$ for our financial document datasets but using $k \in \{10, 15, 20\}$ for MMDocRAG to follow its metrics.

\subsection{Object Detection ($T_{\textit{OD}}$)}
The performance of object detection is measured by the spatial overlap between the predicted bounding box and the ground truth, as well as the inference efficiency.

\textbf{1. Intersection over Union (IoU):} IoU quantifies the overlap between the predicted bounding box $B_p$ and the ground-truth bounding box $B_{gt}$. It is calculated as the area of their intersection divided by the area of their union:
\begin{equation}
\text{IoU} = \frac{\text{Area}(B_p \cap B_{gt})}{\text{Area}(B_p \cup B_{gt})}
\end{equation}
We report the mean IoU across all test samples.

\textbf{2. Inference Time (Time):} We record the total running time required to process the test set and report it as minutes per 100 test cases. This metric highlights the efficiency gains of the task-specific head ($E_{\textit{T}}$) compared to sequential token generation ($E_{\textit{ntp}}$).

\subsection{Anomaly Detection ($T_{\textit{AD}}$)}
For the binary classification-based anomaly detection task, we evaluate both threshold-dependent and threshold-independent metrics to account for the class imbalance.

\textbf{1. Accuracy (Acc):} The ratio of correctly predicted samples (both normal and anomalous) to the total number of samples:
\begin{equation}
\text{Acc} = \frac{\textit{TP} + \textit{TN}}{\textit{TP} + \textit{TN} + \textit{FP} + \textit{FN}}
\end{equation}
where $\textit{TP}, \textit{TN}, \textit{FP}, \textit{FN}$ represent true positives, true negatives, false positives, and false negatives, respectively.

\textbf{2. AUROC:} The Area Under the Receiver Operating Characteristic curve. It plots the True Positive Rate ($\textit{TPR} = \displaystyle\frac{\textit{TP}}{\textit{TP}+\textit{FN}}$) against the False Positive Rate ($\textit{FPR} = \displaystyle\frac{\textit{FP}}{\textit{FP}+\textit{TN}}$) at various threshold settings. It measures the probability that a randomly chosen positive instance is ranked higher than a randomly chosen negative one.

\textbf{3. AUPRC:} The Area Under the Precision-Recall Curve. Precision ($P = \displaystyle\frac{\textit{TP}}{\textit{TP}+\textit{FP}}$) is plotted against Recall ($R = \displaystyle\frac{\textit{TP}}{\textit{TP}+\textit{FN}}$). AUPRC is a more robust metric for the highly imbalanced datasets used in our anomaly detection scenarios, as it does not reward high numbers of True Negatives.

\section{Implementation Details}
\label{sec:more_implementation_details}
We introduce details of the self-annotation process, the fine-tuning settings, baseline settings, and various hyper-parameters for different tasks and different LLMs in this section.

\subsection{Self-Annotation for Information Retrieval ($T_{\textit{IR}}$)}
The self-annotation process includes using the LLM to generate text relevant to the input document or image. In this framework, we prompt the LLM to generate a specific query that acts as a semantic anchor. The prompt for text-to-text retrieval is:

\begin{promptbox}[Prompt for Text-to-text Retrieval Self-Annotation]
Document:\\
<doc>\\
Based on the above document, generate one question that can be answered using the information provided.
\end{promptbox}

The prompt for text-to-image retrieval is:
\begin{promptbox}[Prompt for Text-to-image Retrieval Self-Annotation]
<image title information> \\
Based on the above image, generate one question that can be answered using the information provided.
\end{promptbox}
Since the required annotations are direct, input-related textual queries, no additional post-processing is required for this task.

\subsection{Self-Annotation for Object Detection ($T_{\textit{OD}}$)}

Due to the significant variance in the native capabilities of different models regarding $T_{\textit{OD}}$, we use three distinct self-annotation strategies: Direct Ask, Fuzzy Location, and the OCR-supported method.

\subsubsection{Direct Ask}
This method is applied to Qwen2-VL, as the base model possesses inherent grounding capabilities and can output formalized coordinates. We evaluate this model's potential using both one-turn and multi-turn prompting strategies. The one-turn prompt is:
\begin{promptbox}[Prompt for One-turn Object Detection]
You are given a screenshot of a document page containing a target object: either a chart, table, or figure. Your task is to identify the bounding box of the whole object, including its title, and the text underlying the source of this object. Output normalized coordinates of the bounding box relative to the image size.
\end{promptbox}

The multi-turn prompts decompose the task into identifying the main object, its caption, and its source individually:
\begin{promptbox}[Prompts for Multi-turn Object Detection]
You are given a screenshot of a document page containing a target object: either a chart, table, or figure. Your task is to identify the bounding box of the whole object. Output normalized coordinates of the bounding box relative to the image size.\\

You are given a screenshot of a document page containing a target object: either a chart, table, or figure. Your task is to identify the bounding box of this object's caption Output normalized coordinates of the bounding box relative to the image size.\\

You are given a screenshot of a document page containing a target object: either a chart, table, or figure. Your task is to identify the bounding box of the text underlying the source of this object. Output normalized coordinates of the bounding box relative to the image size.
\end{promptbox}

All outputs will be processed as four numerical numbers to represent the bounding boxes. For one-turn prompts, the output is used directly as the self-annotation. For multi-turn prompts, we calculate the minimum bounding rectangle that encompasses all three predicted boxes to form the final annotation.

\subsubsection{Fuzzy Location}
Since LLaVA-1.6 and Phi-3-VL lack native grounding abilities, we implement a bisection-based post-processing strategy to convert semantic understanding into spatial coordinates. We first prompt the model to describe the object:

\begin{promptbox}[Prompt for Object Description]
You are given a screenshot of a document page which may contain a target object: either a chart, table, or figure. Briefly describe this object in no more than five words.
\end{promptbox}

We apply this query to the original image and store the description. We then iteratively crop the image (e.g., moving the bottom boundary to half the image height) and re-query the model. If the semantic similarity between the new and original description exceeds 60\%, we assume the object remains largely within the crop and continue the bisection. If similarity drops, we backtrack. This process repeats for all four boundaries until the boundary shift is less than 30 pixels, resulting in a "fuzzy" bounding box.

\subsubsection{OCR-supported Method}
To enhance the accuracy of annotations for weaker models, we utilize pytesseract, a classic non-deep-learning OCR package. This method provides coordinate support while remaining computationally much lighter than model-based approaches. We prompt the model to extract surrounding text like:
\begin{promptbox}[Prompt for Surrounding Text Extraction]
You are given a screenshot of a document page containing a target object: either a chart, table, or figure. Output the title of this object. \\
The title of this object is:
\end{promptbox}

We then use the OCR engine to locate the output text on the page. By combining the bounding boxes of the extracted text anchors, we derive the smallest rectangle that covers the target area as the self-annotation.

\subsection{Self-Annotation for Anomaly Detection ($T_{\textit{AD}}$)}
For anomaly detection, we provide the model with an image-query pair and use the following prompt:
\begin{promptbox}[Prompt for Surrounding Text Extraction]
Based on the image above, determine whether the following question can be answered using the information provided in the image. Respond with 'yes' or 'no'. \\
Question: <question>.
\end{promptbox}
The textual responses are subsequently mapped to binary class labels (0 or 1) to form the training set.

\subsection{Fine-tuning Details}
The fine-tuning process uses task-specific output layers and specialized objective functions. 
For $T_{\textit{IR}}$, we bypass the base model’s token prediction layer and use the hidden state of the final token from the last layer as the dense representation. The model is fine-tuned using a standard contrastive learning objective. 
For $T_{\textit{OD}}$, we attach a regression head, a three-layer MLP, to the final hidden state of LLMs. The hidden size of the regression head is 1,024, and the final layer outputs 4 dimensions followed by a Sigmoid function to normalize coordinates between 0 and 1. We use $L_1$ loss for fine-tuning. 
For $T_{\textit{AD}}$, we add a classification head (three-layer MLP, hidden size 1,024). The output is a single logit used for binary classification, optimized via binary cross-entropy loss.

All training sets are split into training and validation data at a 9:1 ratio. We use LoRA to adapt the models, adding trainable parameters to the query, key, value, and output projections ($W_q, W_k, W_v, W_o$) of all attention layers. The rank ($r$) is set to 8 and lora alpha ($\alpha$) to 16. The total trainable parameters account for less than 0.5\% of the total model size.
The training configurations are tailored to the specific requirements of each task. For $T_{\textit{IR}}$, the model is trained for 5 epochs with a batch size of 4, while $T_{\textit{OD}}$ utilizes a batch size of 4 over 20 epochs. In the case of $T_{\textit{AD}}$, the training duration is extended to 50 epochs with a batch size of 4. Across all three tasks, the models are optimized using the Adam optimizer~\cite{kingma2015adam} with a consistent learning rate of $1 \times 10^{-4}$.

\section{Results on Open-source datasets}
\label{sec:open_source_data_results}

\subsection{IR results on MMDocRAG, DocVQA-2020, and SciMMIR}
\label{sec:ir_opensource_results}

\begin{table}[t]
\vspace{-0.0cm}
\renewcommand\arraystretch{1}
\caption{Performance comparison on MMDocRAG's information retrieval task. SKR's results are highlighted in bold. Across all models and modalities, SKR consistently achieves the best performance. Qwen2 models adapted via SKR outperform specialized industrial baselines.}
\vspace{-0.0cm}
\begin{center}
\small
\begin{tabular}[b]{l@{\hspace{3pt}} l r@{\hspace{10pt}} r@{\hspace{10pt}} r@{\hspace{10pt}} }
\toprule
\multicolumn{2}{l}{Model}  & R@10 & R@15 & R@20 \\
\cmidrule{1-5}
\multicolumn{2}{l}{DSE$_{wiki-ss}$}   & 0.671 & 0.753 & 0.799 \\
\multicolumn{2}{l}{DSE$_{docmatix}$}  & 0.657 & 0.750 & 0.782 \\
\multicolumn{2}{l}{ColPali}  & 0.682 & 0.775 & 0.812 \\
\multicolumn{2}{l}{ColQwen}  & 0.708 & 0.792 & 0.843 \\
\cmidrule{1-2}
\multirow{2}{*}{LLaVA-1.6}
 & Base & 0.527 & 0.634 & 0.705 \\
 & \textbf{SKR}  & \textbf{0.834} & \textbf{0.886} & \textbf{0.920} \\
\cmidrule{1-2}
\multirow{2}{*}{Phi-3-VL}
 & Base & 0.607 & 0.703 & 0.768 \\
 & \textbf{SKR}  & \textbf{0.842} & \textbf{0.891} & \textbf{0.923} \\
\cmidrule{1-2}
\multirow{2}{*}{Qwen2-VL}
 & Base & 0.578 & 0.687 & 0.749 \\
 & \textbf{SKR}  & \textbf{0.877} & \textbf{0.916} & \textbf{0.944} \\

\bottomrule
\end{tabular}
\end{center}
\label{tab:mmdocrag_results}
\vspace{-0.4cm}
\end{table}
Table~\ref{tab:mmdocrag_results} presents the performance of SKR on the open-source MMDocRAG benchmark. To ensure a fair comparison, we use the reported results of the four top-performing models in text-to-image retrieval from their original publications. Our experimental setup strictly adheres to their established evaluation protocol and metrics. These baselines are leading retrieval models, including DSE~\cite{ma2024unifying}, built on the Phi-3-Vision architecture, and ColQwen~\cite{faysse2025colpali}, which uses the Qwen2-VL backbone.

The results demonstrate that while specialized fine-tuning (as seen in DSE and ColQwen) provides a performance boost over base MLLMs, SKR consistently outperforms all industrial baselines across the evaluated architectures. Notably, SKR-adapted models achieve at least a 12.6\% absolute improvement in Recall@10 compared to the strongest industrial baseline, ColQwen.

\begin{table}[h]
\vspace{-0.0cm}
\caption{Performance comparison on the DocVQA-2020 and SciMMIR datasets. SKR's results are highlighted in bold.}
\begin{center}
\vspace{-0.0cm}
\begin{tabular}{l c@{\hspace{10pt}} c@{\hspace{10pt}} c@{\hspace{10pt}} c@{\hspace{10pt}} c}
\toprule
\textbf{Model} & \textbf{MRR} & \textbf{R@1} & \textbf{R@3} & \textbf{R@5} & \textbf{R@10} \\ \cmidrule{1-6}
\multicolumn{6}{l}{DocVQA-2020} \\
\cmidrule{1-6}
Nomic & 0.848 & 0.754 & 0.931 & 0.970 & 0.994 \\
\textbf{Qwen-SKR} & \textbf{0.931} & \textbf{0.882} & \textbf{0.976} & \textbf{0.993} & \textbf{0.999} \\ 
\cmidrule{1-6}
\multicolumn{6}{l}{SciMMIR} \\
\cmidrule{1-6}
Nomic & 0.326 & 0.283 & 0.440 & 0.491 & 0.531 \\
\textbf{Qwen-SKR} & \textbf{0.562} & \textbf{0.467} & \textbf{0.622} & \textbf{0.675} & \textbf{0.732} \\ \bottomrule
\end{tabular}
\end{center}
\label{tab:generalization_results}
\vspace{-0.6cm}
\end{table}

Furthermore, Table~\ref{tab:generalization_results} highlights the results of Qwen-SKR (fine-tuned exclusively on MMDocRAG) when evaluated on the DocVQA-2020 and SciMMIR test sets. Despite having no exposure to these datasets during the SKR process, our model significantly outperforms the Nomic baseline. This demonstrates SKR’s exceptional cross-dataset generalization for similar tasks.

The fact that SKR, using only unannotated images, surpasses models trained on massive, annotated retrieval datasets proves that the expression bottleneck is the primary limiting factor in current multi-modal RAG systems. By transitioning to a task-optimized expression mechanism, SKR allows the model to use its internal knowledge far more effectively in specific tasks.

\subsection{OD and AD results on CUB\_200\_2011}
\label{sec:od_ad_opensource_results}
Tables~\ref{tab:opensource_od_results} and \ref{tab:opensource_ad_results} present the performance of SKR on the open-source CUB\_200\_2011 benchmark for OD and AD tasks, respectively. For $T_{\textit{OD}}$, the IoU improved across all models, with the runtime being reduced by at least 88\%. This performance trend on publicly available data directly mirrors the efficiency gains and accuracy improvements observed on our specialized financial dataset.

\begin{table}[b]
\vspace{-0.6cm}
\caption{Performance comparison on $T_{\textit{OD}}$. Results for the SKR method are highlighted in bold. Runtime is reported in minutes per 100 test cases.}
\begin{center}
\label{tab:opensource_od_results}
\begin{tabular}{lcc}
\toprule
Model & IoU & Time \\ 
\midrule
Qwen Base & 0.687 & 2.7 \\
\textbf{Qwen SKR} & \textbf{0.692} & \textbf{0.3} \\ 
\midrule
LLaVA Base & 0.275 & 20.2 \\
\textbf{LLaVA SKR} & \textbf{0.311} & \textbf{0.4} \\ 
\midrule
Phi3 Base & 0.237 & 19.3 \\
\textbf{Phi3 SKR} & \textbf{0.266} & \textbf{0.4} \\ 
\bottomrule
\end{tabular}
\end{center}
\end{table}

For $T_{\textit{AD}}$, consistent with the findings on our financial dataset, accuracy improvements are marginal due to the extreme class imbalance inherent in the task. However, the AUROC and AUPRC scores, which are more diagnostic of anomaly detection performance, show significant gains across all models.

\begin{table}[t]
\caption{Performance comparison on $T_{\textit{AD}}$. SKR results are highlighted in bold. Runtime is reported in minutes per 100 test cases.}
\label{tab:opensource_ad_results}
\begin{center}
\begin{tabular}{lcccc}
\toprule
Model & Acc $\uparrow$ & AUROC & AUPRC & Time \\ \midrule
Qwen Base & 0.985 & 0.857 & 0.471 & 0.8 \\
\textbf{Qwen SKR} & \textbf{0.986} & \textbf{0.870} & \textbf{0.545} & \textbf{0.3} \\ \midrule
LLaVA Base & 0.965 & 0.840 & 0.246 & 2.2 \\
\textbf{LLaVA SKR} & \textbf{0.966} & \textbf{0.918} & \textbf{0.442} & \textbf{0.3} \\ \midrule
Phi3 Base & 0.971 & 0.863 & 0.300 & 2.7 \\
\textbf{Phi3 SKR} & \textbf{0.980} & \textbf{0.882} & \textbf{0.476} & \textbf{0.3} \\ \bottomrule
\end{tabular}
\end{center}
\vspace{-0.4cm}
\end{table}

These results on publicly available data further confirm that SKR  is highly effective in both tasks, significantly improving performance while reducing runtime.

\section{Impact of Model Scaling}
\label{sec:model_size_analysis}

To investigate the relationship between model scale and the effectiveness of SKR, we evaluate the Gemma-3 model family~\cite{team2025gemma} across three parameter scales: 4b, 12b, and 27b. We specifically use the Gemma-3 family for this study rather than the models evaluated in our primary experiments to ensure a controlled comparison. While the previous evaluated models (e.g., Qwen, Phi, LLaVA) exhibit architectural and generational variances that could confound results, the Gemma-3 family provides a homogeneous environment where model size is the primary variable. This allows us to more rigorously isolate the impact of parameter scale on the SKR process.

The results, summarized in Table~\ref{tab:gemma_results}, provide insights into how model size influences both the quality of intrinsic knowledge and the ultimate ceiling for task adaptation. In addition to the standard SKR pipeline, we include two cross-model supervision variants: (1) 27b-ft, where the 4b model is fine-tuned using annotations from the 27b model, and (2) 4b-ft, where the 27b model is fine-tuned using annotations from the 4b model.

\begin{table}[t]
\vspace{-0.0cm}
\renewcommand\arraystretch{1}
\caption{Performance comparison on $T_{\textit{IR}}$ for Gemma-3 models. SKR's results are highlighted in bold. The lines ``27b-ft'' and ``4b-ft'' denote models fine-tuned using annotations generated by the Gemma-3-27b-it and Gemma-3-4b-it models, respectively.}
\vspace{0.1cm}
\begin{center}
\small
\begin{tabular}[b]{l@{\hspace{3pt}} l r@{\hspace{10pt}} r@{\hspace{10pt}} r@{\hspace{10pt}} r@{\hspace{10pt}} r@{\hspace{10pt}}}
\toprule
\multicolumn{2}{l}{Model} & MRR & R@1 & R@3 & R@5 & R@10 \\
\cmidrule{1-2}
\multirow{4}{*}{\shortstack{Gemma-3\\4b-it}}
 & Base & 0.003 & 0.000 & 0.000 & 0.000 & 0.000 \\
 & E5-V & 0.222 & 0.165 & 0.250 & 0.275 & 0.330 \\
 & \textbf{SKR} & \textbf{0.415} & \textbf{0.345} & \textbf{0.415} & \textbf{0.485} & \textbf{0.590} \\
 & 27b-ft & 0.501 & 0.410 & 0.540 & 0.595 & 0.660 \\
\cmidrule{1-2}
\multirow{3}{*}{\shortstack{Gemma-3\\12b-it}}
 & Base & 0.025 & 0.010 & 0.020 & 0.025 & 0.045 \\
 & E5-V & 0.293 & 0.230 & 0.305 & 0.355 & 0.425 \\
 & \textbf{SKR} & \textbf{0.526} & \textbf{0.420} & \textbf{0.580} & \textbf{0.635} & \textbf{0.736} \\
\cmidrule{1-2}
\multirow{4}{*}{\shortstack{Gemma-3\\27b-it}}
 & Base & 0.051 & 0.035 & 0.050 & 0.070 & 0.080 \\
 & E5-V & 0.047 & 0.025 & 0.060 & 0.065 & 0.080 \\
 & \textbf{SKR} & \textbf{0.531} & \textbf{0.440} & \textbf{0.580} & \textbf{0.625} & \textbf{0.715} \\
 & 4b-ft & 0.472 & 0.375 & 0.535 & 0.580 & 0.635 \\

\bottomrule
\end{tabular}
\end{center}
\label{tab:gemma_results}
\vspace{-0.4cm}
\end{table}

Our results indicate that model size positively correlates with both the base model knowledge level and the performance of SKR-adapted models. The 12B and 27B versions outperform the 4B model, suggesting that larger architectures are more adept at internalizing and re-expressing information. However, this benefit exhibits asymptotic behavior; the performance gap between the 12B and 27B models is marginal, indicating a potential upper limit for the SKR process on this specific $T_{\textit{IR}}$ task.

\begin{figure*}[t]
\centering

\subfigure[{
    \parbox[t]{0.25\linewidth}{ Layer-wise CKA similarity across methods for Qwen2-VL.}
}]{
    \label{fig:qwen_sub}
    \includegraphics[width=0.31\linewidth]{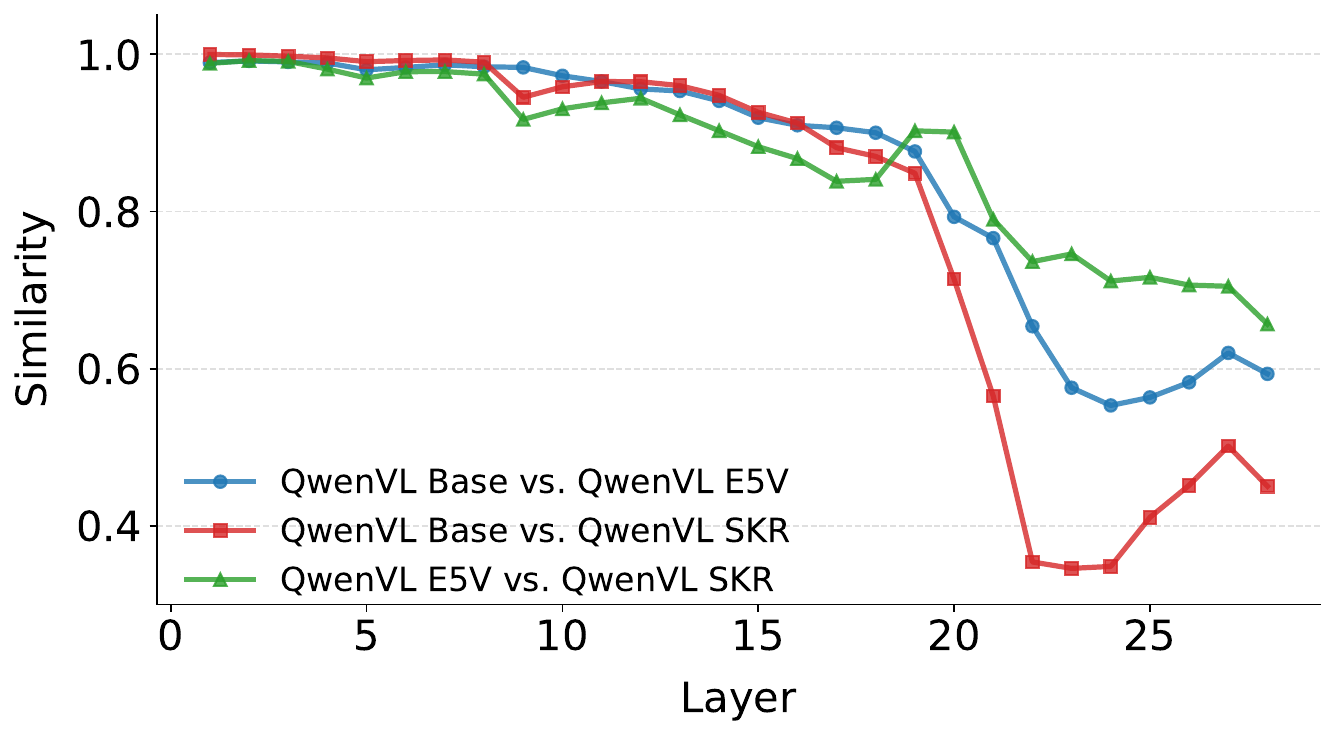}
}
\hfill
\subfigure[{
    \parbox[t]{0.25\linewidth}{ Layer-wise CKA similarity across methods for LLaVA-1.6.}
}]{
    \label{fig:llava_sub}
    \includegraphics[width=0.31\linewidth]{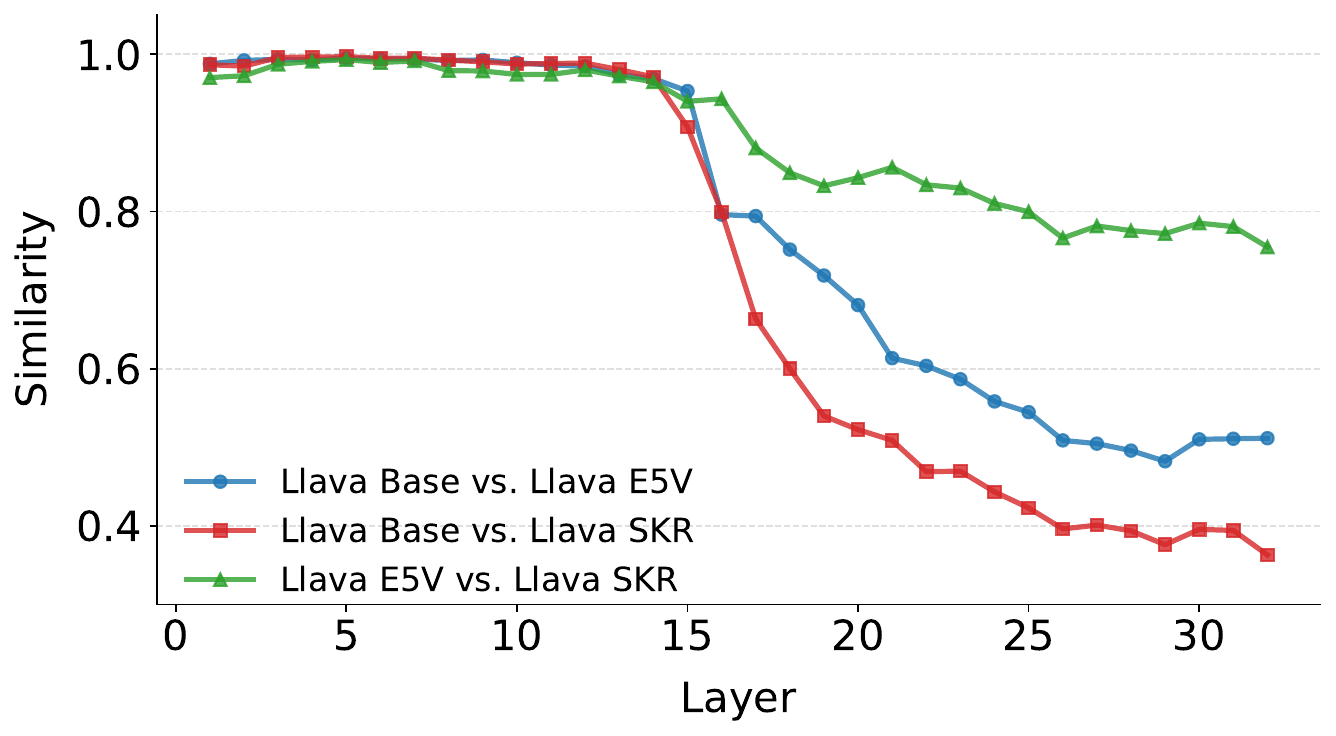}
}
\hfill
\subfigure[{
    \parbox[t]{0.25\linewidth}{ Layer-wise CKA similarity between Qwen2-VL and LLaVA-1.6 under the same method.}
}]{
    \label{fig:combined_sub}
    \includegraphics[width=0.31\linewidth]{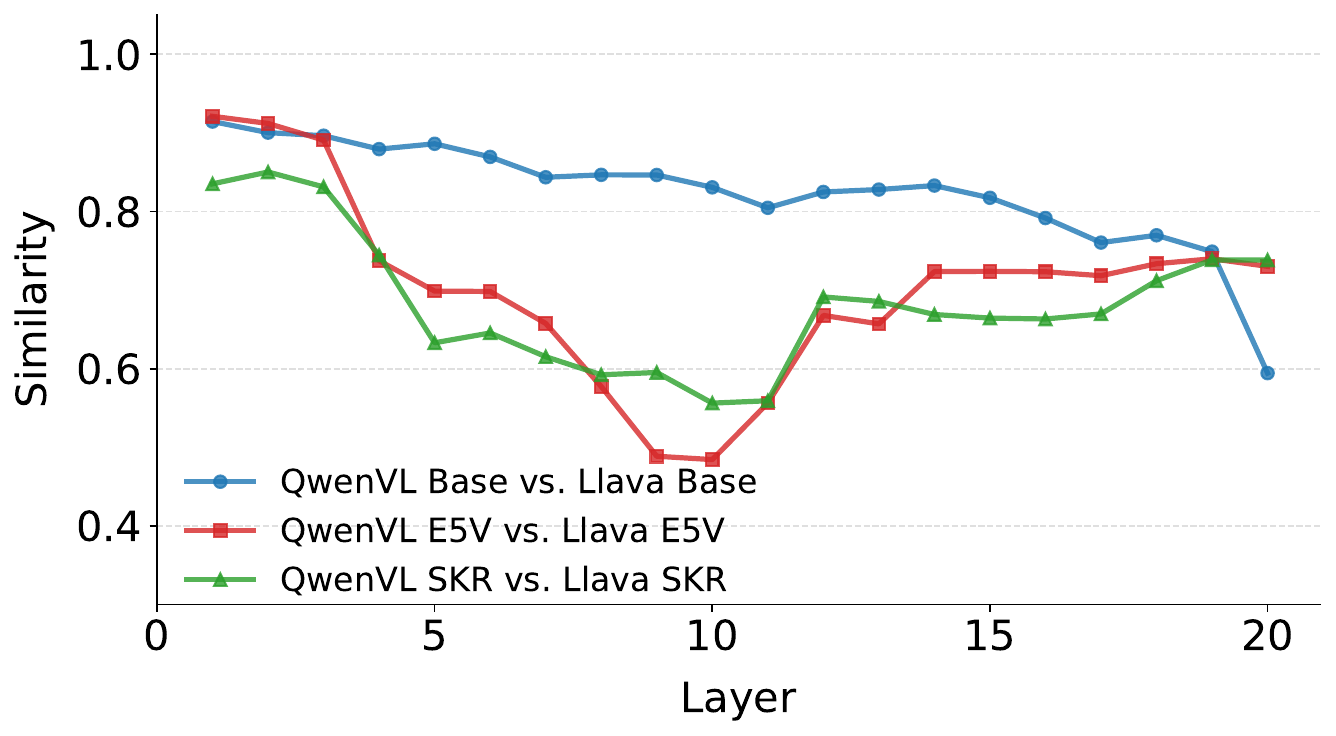}
}

\caption{Layer-wise representation similarity (CKA) under different settings.}
\label{fig:total_comparison}
\end{figure*}

The performance gains observed in larger models can be attributed to two distinct factors: annotation quality (the quality of self-generated annotations) and expressive capacity (the model's inherent ability to be adapted). The cross-model supervision experiments provide a clear decoupling of these factors:

The 4b model's performance increases significantly when fine-tuned on 27b-generated labels (27b-ft) compared to its own self-annotations. Conversely, the 27b model's performance drops when using 4b-generated labels (4b-ft). This confirms that larger models possess superior intrinsic knowledge, leading to higher-quality self-annotation.

Even when provided with high-quality 27b annotations, the 4b model still fails to match the performance of the 12b or 27b SKR-adapted models. Furthermore, the 27b model fine-tuned on 4b annotations (4b-ft) still outperforms the 4b model's best SKR results. This suggests that larger models have a higher architectural ceiling for re-expression, a more potential latent space that responds better to task-specific adaptation.

In summary, model size affects both the model's intrinsic knowledge and its potential for re-expression. Larger models not only generate superior self-annotations but also offer more potential for task-specific improvement. However, these gains do not scale indefinitely. Our findings suggest that we should calibrate model size to task difficulty, selecting the most efficient architecture that meets the required performance threshold.

\section{Ablation Study on LoRA Adapters}
\label{sec:lora_ablation}

The SKR framework is designed as a knowledge re-expression process rather than a mere formatting alignment for specific tasks. The LoRA adapters are essential to harmonize the model’s internal parametric knowledge with the newly adapted output mechanism. To validate this design, we conducted ablation experiments comparing the full SKR framework against a "Head-only" baseline. In the Head-only configuration, we implement the task-specific head and loss functions but keep the backbone parameters frozen, bypassing the parameter adaptation phase.

Tables~\ref{tab:lora_abaltion_ir}, \ref{tab:lora_abaltion_od}, and \ref{tab:lora_abaltion_ad} summarize these comparative results across retrieval, detection, and anomaly detection tasks.

\begin{table}[t]
\centering
\caption{Performance comparison on $T_{\textit{IR}}$ on the MMDocRAG dataset.}
\begin{center}
\label{tab:lora_abaltion_ir}
\begin{tabular}{lccc}
\toprule
Model & R@10 & R@15 & R@20 \\ \midrule
Qwen Base & 0.578 & 0.687 & 0.749 \\
Qwen Head-only & 0.652 & 0.735 & 0.797 \\
\textbf{Qwen SKR} & \textbf{0.877} & \textbf{0.916} & \textbf{0.944} \\ \bottomrule
\end{tabular}
\end{center}
\vspace{-0.6cm}
\end{table}

\begin{table}[b]
\centering
\vspace{-0.6cm}
\caption{Performance comparison on $T_{\textit{OD}}$ on the CUB dataset.}
\begin{center}
\label{tab:lora_abaltion_od}
\begin{tabular}{lc}
\toprule
Model & IoU \\ \midrule
Qwen Base & 0.687 \\
Qwen Head-only & 0.491 \\
\textbf{Qwen SKR} & \textbf{0.692} \\ \bottomrule
\end{tabular}
\end{center}
\end{table}

\begin{table}[t]
\centering
\caption{Performance comparison on $T_{\textit{AD}}$ on the CUB dataset.}
\begin{center}
\label{tab:lora_abaltion_ad}
\begin{tabular}{lccc}
\toprule
Model & Acc & AUROC & AUPRC \\ \midrule
Qwen Base & 0.985 & 0.857 & 0.471 \\
Qwen Head-only & 0.980 & 0.856 & 0.492 \\
\textbf{Qwen SKR} & \textbf{0.986} & \textbf{0.870} & \textbf{0.545} \\ \bottomrule
\end{tabular}
\end{center}
\vspace{-0.6cm}
\end{table}

The results clearly demonstrate that modifying the output head alone is insufficient for effective task adaptation. In $T_{\textit{IR}}$, the R@10 gain achieved by SKR (0.299) is over 400\% greater than that of the Head-only baseline (0.074). In $T_{\textit{OD}}$, the Head-only approach significantly degrades performance, with IoU dropping to 0.491, whereas SKR maintains and slightly improves upon the base model's accuracy while dramatically increasing efficiency. In $T_{\textit{AD}}$, while the Head-only model shows a marginal increase in AUPRC (0.021), it remains substantially lower than the improvement provided by SKR (0.074).

Collectively, these findings confirm that our performance gains are not a trivial byproduct of using task heads. Instead, they stem from the SKR's unique ability to effectively re-align latent representations with a new expression paradigm. This proves that for successful zero-shot adaptation, internal knowledge re-expression is a fundamental necessity.

\section{CKA Analysis}
\label{sec:cka_analysis}

We analyze representation similarity across layers, methods, and architectures to gain deeper insights into how task-specific capabilities emerge within the model's latent space.

We use Center Kernel Alignment (CKA~\cite{kornblith2019similarity}) to quantify these similarities. CKA measures the alignment between representations in a kernel space, providing a robust metric that is invariant to isotropic scaling and orthogonal transformations. This method is widely adopted for studying the internal representational characteristics of LLMs~\cite{zhao2024layer}.

Figures \ref{fig:qwen_sub} and \ref{fig:llava_sub} illustrate the layer-wise CKA similarity for Qwen2-VL and LLaVA-1.6 when comparing different adaptation methods. Several key patterns emerge. In both architectures, representations remain highly similar in the early layers but diverge significantly as the data progresses into deeper layers. The degree of divergence is closely tied to the performance gap between methods. Because the SKR-adapted models achieve the highest performance, their similarity to the original base models is the lowest, dropping sharply after the midpoint layers. Instead, the similarity between E5-V and SKR (green lines) remains relatively high compared to their similarity to the Base model. This suggests that for a given model, different task-adaptation methods shift internal representations in a consistent direction toward task-specific utility. The dramatic drop in similarity in the deepest layers, those immediately preceding the output head, confirms that SKR fundamentally reconfigures the model's outputs to overcome the default NTP bottleneck.

Then Figure~\ref{fig:combined_sub} provides a joint comparison between the two different base architectures (Qwen2-VL vs. LLaVA-1.6). As the two models differ in depth, we align them by selecting the last 20 layers for comparison. Even when both models are adapted for the same task using the same SKR method, their representational similarity consistently remains below 0.8. While both E5-V and SKR adaptation drive the models toward a shared task-oriented objective, they do not erase the fundamental difference of the base model's parametric knowledge. This underscores the core premise of SKR: it optimizes the expression of a model's existing intrinsic knowledge rather than forcing diverse models into a single, homogenized state. The models reach the same high-performance goal while maintaining their unique internal representational knowledge.

\end{document}